\pdfoutput=1

\documentclass[11pt]{article}

\usepackage[]{emnlp2021}

\usepackage{times}
\usepackage{latexsym}

\usepackage[T1]{fontenc}

\usepackage[utf8]{inputenc}

\usepackage[letterspace=-26]{microtype}

\usepackage{amsmath}
\usepackage{amssymb}
\usepackage{mathtools}
\usepackage{arydshln}
\usepackage{booktabs,colortbl}
\usepackage{tabularx}
\usepackage{multirow}
\usepackage{tikz}
\usepackage{pgfplots}
\usepackage{xcolor}
\usepackage[normalem]{ulem}
\usepackage{dashbox}%
\usepackage{textcomp}
\usepackage{tcolorbox}
\usepackage{adjustbox}
\usepackage{lipsum}
\usepackage[inline]{enumitem}
\usepackage{menukeys}
\usepackage{pmboxdraw}
\usepackage{newunicodechar}
\usepackage{graphicx}
\usepackage[shortcuts]{extdash}
\usepackage{pifont}

\pgfplotsset{compat=1.13}

\definecolor{c0}{cmyk}{1,0.3968,0,0.2588} 
\definecolor{c1}{cmyk}{0,0.6175,0.8848,0.1490} 
\definecolor{c2}{cmyk}{0.3081,0,0.7209,0.3255}
\definecolor{c3}{cmyk}{0.1127,0.6690,0,0.4431} 
\definecolor{c4}{cmyk}{0.6765,0.2017,0,0.0667} 
\definecolor{c5}{cmyk}{0,0.8765,0.7099,0.3647} 

\definecolor{darkgrey}{RGB}{149,149,149}
\definecolor{decentgrey}{RGB}{212,212,212}

\def\cca#1#2{\cellcolor{c0!#1}\ifnum #1>43\color{white}\fi\sffamily\scriptsize{#2}}

\usetikzlibrary{calc,fit,positioning,arrows,arrows.meta,backgrounds,decorations.pathreplacing}
\pgfdeclarelayer{bg}
\pgfsetlayers{bg,main}

\newtcbox{\hlprimary}{on line,colback=c0!10,colframe=white,size=fbox,arc=3pt, box align=base,before upper=\strut, top=-2pt, bottom=-4pt, left=-1pt, right=-1pt, boxrule=0pt}
\newtcbox{\hlprimarytab}{on line, box align=base, colback=c0!10,colframe=white,size=fbox,arc=3pt, before upper=\strut, top=-2pt, bottom=-4pt, left=-2pt, right=-2pt, boxrule=0pt}
\newtcbox{\hlsecondary}{on line,colback=c1!10,colframe=white,size=fbox,arc=3pt, box align=base,before upper=\strut, top=-2pt, bottom=-4pt, left=-1pt, right=-1pt, boxrule=0pt}
\newtcbox{\hlsecondarytab}{on line, box align=base, colback=c1!10,colframe=white,size=fbox,arc=3pt, before upper=\strut, top=-2pt, bottom=-4pt, left=-2pt, right=-2pt, boxrule=0pt}
\newtcolorbox{hlmultiline}{on line,colback=decentgrey!75,colframe=white,size=fbox,arc=3pt, box align=base, top=0pt, bottom=2pt, boxrule=0pt, before=\adjustbox{valign=c}\bgroup, after=\egroup, before upper=\strut}

\newcolumntype{Y}{>{\centering\arraybackslash}X}
\newcolumntype{Z}{>{\raggedleft\arraybackslash}X}

\newunicodechar{🦕}{\includegraphics[height=1.25\fontcharht\font`A]{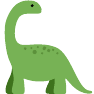}}

\newcommand{\cmark}{{\color{c2}\ding{51}}}%
\newcommand{\xmark}{{\color{c1}\ding{55}}}%

\newcommand\ours{\textsc{Dino}}

\newcommand\oursFull{Datasets from Instructions}
\newcommand\oursFullHl{\textbf{D}atasets from \textbf{In}structi\textbf{o}ns}
\newcommand\ourDs{STS\=/🦕}
\newcommand\ourDsFull{STS\=/🦕\=/$\mathbf{x}_1\mathbf{x}_2$}
\newcommand\ourDsSemi{STS\=/🦕\=/$\mathbf{x}_2$}

\newcommand\given{\,{\mid}\,}
\newcommand{\bt}{\fontseries{b}\selectfont}

\title{Generating Datasets with Pretrained Language Models}

\author{
	Timo Schick \and Hinrich Sch\"utze \\
	Center for Information and Language Processing \\ LMU Munich, Germany \\
	{\tt schickt@cis.lmu.de}
}

\newcounter{notecounter}

\newcommand{\enoteson}{\long\gdef\enote##1##2{{
\stepcounter{notecounter}
{\large\bf
\hspace{1cm}\arabic{notecounter} $<<<$ ##1: ##2
$>>>$\hspace{1cm}}}}}

\enoteson

\begin{document}
	\maketitle
	\begin{abstract}
		To obtain high-quality sentence embeddings
	from pretrained language models (PLMs), they must either be
	augmented with additional pretraining objectives or
	finetuned on a large set of labeled text pairs. While the latter approach typically outperforms the former, it requires great human effort to generate suitable datasets of sufficient size.
		In this paper, we show how
                PLMs can be leveraged to obtain high-quality
	sentence embeddings without the need for labeled data, finetuning or modifications to the
pretraining objective: We utilize the generative abilities
of large and high-performing PLMs to generate entire
	datasets of labeled text pairs from scratch, which
	we then use for finetuning much smaller and more
	efficient models. Our fully unsupervised approach
	outperforms strong  baselines on several semantic
	textual similarity datasets.\footnote{Our
	code and datasets are publicly available at \url{https://github.com/timoschick/dino}.}
	\end{abstract}

\section{Introduction}

While pretrained language models (PLMs) achieve strong results for many NLP tasks
\citep{peters2018deep,radford2018improving,devlin2018bert},
they do not produce good sentence embeddings out of the
box \citep{reimers-gurevych-2019-sentence}. Recent
approaches address this by augmenting or replacing the
language modeling objective with likewise unsupervised
sentence-level
objectives \citep[e.g.,][]{zhang-etal-2020-unsupervised,li-etal-2020-sentence},
but they typically lag behind their supervised counterparts
trained on human-annotated sentence pairs. Unfortunately,
obtaining large amounts of high-quality training data can be
both difficult and prohibitively
expensive \citep{bowman-etal-2015-large,agirre-etal-2016-semeval}. Furthermore,
with larger and larger model sizes \citep{radford2018language,raffel2019exploring,brown2020language,fedus2021switch}, it becomes increasingly challenging to finetune PLMs.

\begin{figure}
	\tikzset{
		instruction/.style={
			thick, font=\sffamily\small\lsstyle, text width=0.9347\linewidth, align=left, draw=decentgrey, inner xsep=6pt, inner ysep=8pt, rounded corners=2pt,
		},
	}
	\centering
	\begin{tikzpicture}
	
	\node[instruction](instruction-0){
		\textbf{Task}:\,Write two sentences that \textcolor{c0}{mean the same thing}. \\[0.2cm]
		\textbf{Sentence 1}: ``\textcolor{c2}{A man is playing a flute.}'' \\[0.2cm]
		\textbf{Sentence 2}: ``\textcolor{c1}{He's playing a flute.''}
	};
	
	\node[instruction, below=0.2cm of instruction-0](instruction-1){
		\textbf{Task}:\,Write two sentences that \textcolor{c0}{are somewhat similar}. \\[0.2cm]
		\textbf{Sentence 1}: ``\textcolor{c2}{A man is playing a flute.}'' \\[0.2cm]
		\textbf{Sentence 2}: ``\textcolor{c1}{A woman has been playing the violin.''}
	};
	
	\node[instruction, below=0.2cm of instruction-1](instruction-2){
		\textbf{Task}: Write two sentences that \textcolor{c0}{are on completely different topics}. \\[0.2cm]
		\textbf{Sentence 1}: ``\textcolor{c2}{A man is playing a flute.}'' \\[0.2cm]
		\textbf{Sentence 2}: ``\textcolor{c1}{A woman is walking down the street.''}
	};	
	\end{tikzpicture}
	\caption{\textcolor{c1}{Continuations} generated by GPT2-XL with \ours{} for three different \textcolor{c0}{task descriptions}. 
We investigate two different unsupervised approaches to
generating sentence-similarity datasets: (i) 
The \textcolor{c2}{input sentence} is given and only
the \textcolor{c1}{continuation} is generated. This requires
that an (unlabeled) set of sentences is
available. (ii) Both \textcolor{c2}{input sentence}
and \textcolor{c1}{continuation} are generated. This does
not rely on the availability of any resources.}	\label{figure:motivational-example}
\end{figure}

To alleviate both problems, we explore a
novel 
approach to obtaining high-quality sentence embeddings:
We 
mimic
the creation of NLI datasets  by
human
crowdworkers \citep{bowman-etal-2015-large,williams2018mnli},
but replace human annotators
with large PLMs. This allows us to automatically create
entire datasets from scratch that can be used for supervised
training of much smaller models. Not only does this solve
the problem of limited training data, it also provides a
viable path to leverage big models like
GPT-3 \citep{brown2020language} without requiring any
updates to their parameters. As illustrated in
Figure~\ref{figure:motivational-example}, our approach is
based on recent methods for providing instructions to
PLMs \citep[e.g.,][]{radford2018language,brown2020language,schick2020fewshot,schick2020exploiting}. We
use the \emph{self-debiasing} approach
of \citet{schick2021selfdiagnosis} to ensure that each
generated text pair is not only a good fit for a given
similarity label, but also \emph{not} a good fit for other
labels. We refer to our method as \oursFullHl{} (\ours{}).

In summary, our contributions are as follows:
\begin{itemize} 
\item We introduce \ours{}, a method for automatically generating labeled datasets of arbitrary size by providing PLMs with instructions.
\item We release \ourDs{} (read as ``STS-Dino''), the
first textual similarity dataset generated
completely automatically,
without any human annotation effort.
\item We show that Sentence-RoBERTa \citep{reimers-gurevych-2019-sentence} trained on \ourDs{} outperforms strong baselines on several semantic textual similarity datasets.
\end{itemize}

\section{Related Work}

There are many unsupervised approaches to obtaining sentence
embeddings, for example by averaging word
embeddings \citep{mikolov2013word2vec,Glove,bojanowski2017enriching}
or with carefully designed sentence-level
objectives \citep{pmlr-v32-le14,NIPS2015_f442d33f}.
Ensembling several methods improves results
\citep{porner19domainadapt,porner20meta}.
Recent work obtains sentence representations by supplementing BERT \citep{devlin2018bert} or other PLMs with additional unsupervised objectives \citep{zhang-etal-2020-unsupervised,li-etal-2020-sentence,wu2020clear,giorgi2020declutr}. Often, labeled datasets such as paraphrase databases \citep{wieting-gimpel-2018-paranmt} or natural language inference datasets \citep{conneau-etal-2017-supervised,cer-etal-2018-universal,reimers-gurevych-2019-sentence} are used for supervised learning.

Some approaches \emph{augment} existing datasets with
automatically generated
examples \citep{Anaby-Tavor_Carmeli_Goldbraich_Kantor_Kour_Shlomov_Tepper_Zwerdling_2020,papanikolaou2020dare,yang-etal-2020-generative,mohapatra2020simulated,kumar2021data},
but in contrast to our work, all of these approaches require
that there already exists a labeled dataset for finetuning the generator.
Providing PLMs with task descriptions for zero- or few-shot learning has been studied extensively \citep[e.g.,][]{radford2018language,puri2019zeroshot,brown2020language,schick2020fewshot,schick2020just,schick2020exploiting,Weller_2020,gao2020making,tam2021improving}. However, none of these approaches is suitable for generating sentence embeddings.

Closely related to our work, \citet{efrat2020turking}
examine the ability of PLMs to follow natural language
instructions for generating examples in place of human
crowdworkers, but
find that their
approach  performs poorly. 

\section{\oursFull{}}
\label{section:method}

Let $M$ be a PLM with vocabulary $V$, $X = V^*$ the set of all token sequences and $Y$  a finite set of semantic similarity labels. Our aim is to generate a dataset $Z \subset X \times X \times Y$ of text \emph{pairs} $(\mathbf{x}_1, \mathbf{x}_2)$ with corresponding similarity labels $y$. For $x \in V$ and $\mathbf{x} \in X$, we denote with $p_M(x \given \mathbf{x})$ the probability that $M$ assigns to $x$ as a continuation of $\mathbf{x}$.

\begin{figure}
	\tikzset{
		instruction/.style={
			thick, font=\sffamily\small, text width=0.9347\linewidth, align=left, draw=decentgrey, inner xsep=6pt, inner ysep=8pt, rounded corners=2pt,
		},
	}
	\centering
	\begin{tikzpicture}
	
	\node[instruction](instruction-0){
		\textbf{Task}:\,Write two sentences that \textcolor{c0}{\normalsize$i_y$}. \\[0.2cm]
		\textbf{Sentence 1}: ``\textcolor{c2}{\normalsize$\mathbf{x}_1$}'' \\[0.2cm]
		\textbf{Sentence 2}: ``
	};
	\end{tikzpicture}
	\caption{Instruction template $I_y(\mathbf{x}_1)$
		for similarity label $y$ and input sentence
		$\mathbf{x}_1$; $i_y$ is described in
		Section~\ref{section:method}. See
		Figure~\ref{figure:motivational-example} for
		three instantiations of the template.}
	\label{figure:instruction-template}
\end{figure}

We first assume that \textbf{we already have access to a set
$X_1 \subset X$ of texts} (e.g., a set of sentences that are
typical of the domain of interest).
This is a realistic setting for many real-world
applications, where large amounts of unlabeled text are
abundant, but it is difficult to obtain \emph{interesting}
and (for our task) \emph{useful}
text pairs and labels. \ours{} requires a set
of \emph{instructions} $\mathcal{I} = \{I_y \mid y \in Y \}$
where each $I_y \in \mathcal{I}$ is a function that, given
an input $\mathbf{x}_1 \in X_1$, prompts its recipient to
generate an appropriate second text $\mathbf{x}_2$. We use
the instruction template in Figure~\ref{figure:instruction-template} and consider three levels of similarity ($Y = \{0, 0.5, 1 \}$), where
\begin{align*}
i_y = \begin{cases} \text{\sffamily\small mean the same thing} & \text{if } y\,{=}\,1\\
\text{\sffamily\small are somewhat similar} & \text{if } y\,{=}\,0.5\\
\text{\sffamily\small are on completely different topics} & \text{if } y\,{=}\,0
\end{cases}
\end{align*}
is loosely based on \citet{cer-etal-2017-semeval}'s
five-level similarity scheme. Note that for all $y$, $I_y$
ends with an opening quotation mark, which
allows us to treat the 
first
quotation mark generated by the PLM as a sign that it
is done.

For a given $\mathbf{x}_1 \in X_1$ and $y \in Y$, we could
directly use the  instructions
$I_y$
to obtain $\mathbf{x}_2$ by continuously sampling tokens
\[
x_k \sim p_M( x_k \given I_y(\mathbf{x}_1), x_1, \ldots, x_{k-1})
\]
starting from $k=1$ until $x_k$ is a quotation mark and
setting $\mathbf{x}_2 = x_1, \ldots, x_{k-1}$. However, we
may want the PLM to generate a text $\mathbf{x}_2$ that is
not only a good fit for instruction $I_y(\mathbf{x}_1)$, but
also
\emph{not} a good fit for some other
instruction $I_{y'}(\mathbf{x}_1)$. We refer to
$y'$ as a \emph{counterlabel} for $y$
and denote the set of $y$'s counterlabels  as
$\text{CL}(y)$. For example, $1 \in \text{CL}(0.5)$ means that
for $y = 0.5$, we want $M$ to generate a sentence
$\mathbf{x}_2$ that is similar to ($y = 0.5$), but at
the same time does \emph{not}
have the same meaning as ($y= 1$) sentence
$\mathbf{x}_1$. We achieve this using
\citet{schick2021selfdiagnosis}'s self-debiasing algorithm:
When sampling the token $x_k$, we consider not just $p_y =
p_M( x_k \given I_y(\mathbf{x}_1), x_1, \ldots, x_{k-1})$
[$x_k$'s  probability given $I_y(\mathbf{x}_1)$], but also
$p_{y'}$ [$x_k$'s probability given $I_{y'}(\mathbf{x}_1)$], for all $y' \in \text{CL}(y)$. We penalize each token $x_k$ for which $p_y$ is lower than \emph{any} $p_{y'}$ by multiplying its probability with a factor
$\alpha = \exp(\lambda \cdot \delta_y)
$ where 
\[
\delta_y = p_y - \max_{y' \in \text{CL}(y)} p_{y'}
\] is the difference between $x_k$'s probability given $I_y(\mathbf{x}_1)$ and its maximum probability given $I_{y'}(\mathbf{x}_1)$ for any $y' \in \text{CL}(y)$, and the \emph{decay constant} $\lambda$ is a hyperparameter.

For settings where \textbf{no set of unlabeled texts $X_1$ is
available}, a straightforward approach would be to use the
phrase shown in Figure~\ref{figure:instruction-template} up
to and including the first quotation mark as an instruction
to let the PLM generate both $\mathbf{x}_1$ and
$\mathbf{x}_2$. However, this approach has at least two
issues: First, generated texts may not match the required schema
(e.g., the
model may never produce the string ``Sentence 2:''). Second,
the set of texts $\mathbf{x}_1$ should ideally be highly
diverse, whereas we want to give the model less leeway when
generating $\mathbf{x}_2$, so
we may want to use different sampling strategies for
$\mathbf{x}_1$ and $\mathbf{x}_2$.

We solve both problems as follows: We first use $I_y$
(Figure~\ref{figure:instruction-template}) up to and
including the first quotation mark
(the one right after ``Sentence 1:'')
to generate
$\mathbf{x}_1$; we stop as soon as the model produces a
quotation mark.
We run this procedure repeatedly until we have a sufficient
number of sentences. These are gathered into
a set $X_1$ and then we proceed exactly as in the case where $X_1$
is already given.

\section{Experiments}

\begin{table*}
	\small
	\renewcommand{\arraystretch}{0.92}
	\begin{tabularx}{\linewidth}{llcYYYYYYYY}
		\toprule
		& \textbf{Model} & \textbf{UD} & \textbf{STS12} & \textbf{STS13} & \textbf{STS14} & \textbf{STS15} & \textbf{STS16} & \textbf{STSb} & \textbf{SICK} & \textbf{Avg.} \\
		\midrule
		\multirow{4}{*}{\rotatebox[origin=c]{90}{sup.}}
		& InferSent, Glove & -- & 52.86 & 66.75 & 62.15 & 72.77 & 66.87 & 68.03 & 65.65 & 65.01 \\
		& USE & -- & 64.49 & 67.80 & 64.61 & 76.83 & 73.18 & 74.92 & 76.69 & 71.22 \\
		& S-BERT (base) & -- & 70.97 & 76.53 & \underline{73.19} & 79.09 & 74.30 & 77.03 & 72.91 & 74.89 \\
		& S-RoBERTa (base) & -- & \underline{71.54} & 72.49 & 70.80 & 78.74 & 73.69 & 77.77 & \underline{74.46}  & 74.21 \\
		\midrule
		\multirow{8}{*}{\rotatebox[origin=c]{90}{unsup.}}
		& Avg. GloVe & -- & 55.14 & 70.66 & 59.73 & 68.25 & 63.66 & 58.02 & 53.76 & 61.32 \\
		& Avg. BERT & -- & 38.78 & 57.98 & 57.98 & 63.15 & 61.06 & 46.35 & 58.40 & 54.81 \\
		& BERT CLS & -- & 20.16 & 30.01 & 20.09 & 36.88 & 38.08 & 16.50 & 42.63 & 29.19 \\	
		& \citet{zhang-etal-2020-unsupervised} & NLI & 56.77 & 69.24 & 61.21 & 75.23 & 70.16 & 69.21 & 64.25 & 66.58 \\
		& \citet{li-etal-2020-sentence} & NLI & 59.54 & 64.69 & 64.66 & 72.92 & 71.84 & 58.56 & 65.44 & 65.38 \\
		& \citet{li-etal-2020-sentence} & STS & 63.48 & 72.14 & 68.42 & 73.77 & 75.37 & 70.72 & 63.11 & 69.57 \\
		& \ours{} (\ourDsFull) & -- & 64.87 & 78.30 & 66.38 & 79.60 & 76.47 & 76.51 & \textbf{74.26} & 73.77 \\
		& \ours{} (\ourDsSemi) & STS & \textbf{70.27} & \textbf{\underline{81.26}} & \textbf{71.25} & \textbf{\underline{80.49}} & \textbf{\underline{77.18}} & \textbf{\underline{77.82}} & 68.09 & \textbf{\underline{75.20}} \\
		\bottomrule
	\end{tabularx}
	\caption{Spearman's rank correlation on STS12--16,
	STSb and SICK without finetuning on task-specific examples for
	models with NLI supervision (``sup.'') and fully
	unsupervised (``unsup.'') models using the same
	evaluation setup
	as \citet{reimers-gurevych-2019-sentence}. The
	second column shows which unlabeled data (``UD'') is
	used by unsupervised approaches in addition to
	original pretraining data; the final column shows
	average performance. Results for all baselines
	except \citet{zhang-etal-2020-unsupervised}
	and \citet{li-etal-2020-sentence} are
	from \citet{reimers-gurevych-2019-sentence}. The
	best unsupervised result is shown in bold, the best
	overall result is underlined.
\ours{}  outperforms all unsupervised approaches and,
	surprisingly, also supervised approaches on
four out of six
STS datasets.}
	\label{table:main-results}
\end{table*}

We evaluate \ours{} on several English semantic
textual similarity datasets: the STS tasks 2012--2016 \citep{10.5555/2387636.2387697,agirre-etal-2013-sem,agirre-etal-2014-semeval,agirre-etal-2015-semeval,agirre-etal-2016-semeval},
the STS benchmark (STSb) \citep{cer-etal-2017-semeval}, and
the SICK-Relatedness dataset
(SICK) \citep{marelli-etal-2014-sick}. For all tasks, we
adopt the unsupervised setting without task-specific training examples.

We use \ours{} to generate \ourDs{} ${\subset}\,X \times X \times Y$, a dataset
of text pairs with semantic similarity labels. We generate two variants:
\begin{itemize}[topsep=0.5em]
	\setlength\itemsep{-0.1em}
	\item \ourDsSemi{}, for which we make use of STSb to obtain a set of
	texts $X_1$;
	\item \ourDsFull{}, where the set of sentences $X_1$ is generated from scratch.
\end{itemize}
We use GPT2-XL as PLM with
a decay constant of $\lambda = 100$
and the
set of counterlabels
$\text{CL}(y) = \{ y' \in
Y \mid y' > y \}$. That is, we do not restrict the PLM when generating texts for $y=1$, but for $y=0.5$ ($y=0$) we encourage it not to generate texts $\mathbf{x}_2$ that mean the same thing as (are somewhat similar to) $\mathbf{x}_1$. We apply
top-$p$ \citep{Holtzman2020The} and
top-$k$ \citep{fan-etal-2018-hierarchical,holtzman-etal-2018-learning}
sampling with $p = 0.9$, $k = 5$ and generate up to 40
output tokens. For each $\mathbf{x}_1 \in X_1$ and $y \in
Y$, we generate up to two corresponding
$\mathbf{x}_2$'s.\footnote{As the PLM may not 
generate a quotation mark in the first 40
tokens, we use up to 5 tries to generate the two
$\mathbf{x}_2$'s.}
For \ourDsFull{}, we obtain $X_1$ by generating 15,000 sentences using only top-$p$ sampling (again with $p = 0.9$) and no top-$k$ sampling to ensure more diversity in the generated output.
We remove all examples where $\mathbf{x}_1 = \mathbf{x}_2$ (as those provide no training signal to the model) and 
split the datasets 90/10 into training and validation.

To assess the quality of the generated datasets, we use them to train Sentence-RoBERTa \citep{reimers-gurevych-2019-sentence}, a biencoder architecture based on RoBERTa (base) \citep{liu2019roberta} that measures the similarity of two texts by computing the cosine similarity of their embeddings. 
As our datasets contain many noisy examples, we use a technique similar to label smoothing \citep{szegedy2016rethinking} and replace similarity scores of $0$ and $1$ with $0.1$ and $0.9$, respectively.
Additionally, for each $\mathbf{x}_1$, we sample two
$\mathbf{x}_2$'s from \emph{other} dataset entries and
augment the dataset with $(\mathbf{x}_1, \mathbf{x}_2,
0)$. We use the default parameters
of \citet{reimers-gurevych-2019-sentence} with a batch size
of 32 and train for at most one epoch; the exact number of
training steps is determined based on Spearman's rank correlation on the \ourDs{} validation set.

\paragraph{Results}

\begin{table}
	\small
	\renewcommand{\arraystretch}{0.92}
	\setlength\tabcolsep{3pt}
	\begin{tabularx}{\linewidth}{lcYY}
		\toprule
		\textbf{Model} & \textbf{STS12-16} & \textbf{STSb} & \textbf{SICK} \\
		\midrule
		\ours{} (\ourDsSemi) & \textbf{76.09} & \textbf{77.82} & \textbf{68.09} \\
		\ └ decay constant $\lambda = 0$   & 65.50 & 70.71 & 67.60 \\
		\ └ decay constant $\lambda = 200$ & 75.40 & 77.49 & 66.83 \\
		\ └ no label smoothing & 74.50 & 76.26 & 66.23 \\
		\ └ no augmentation & 70.90 & 73.81 & 63.98 \\
		\bottomrule
	\end{tabularx}
	\caption{Effect of removing self-debiasing ($\lambda = 0$) or increasing the decay constant ($\lambda = 200$), using no label smoothing and performing no data augmentation (sampling random $\textbf{x}_2$'s for each $\textbf{x}_1$) on the performance of \ours{} on STS12-16 (avg), STSb and SICK}
	\label{table:self-debiasing}
\end{table}

We compare S-RoBERTa (base) trained on datasets generated
with \ours{} to S-BERT and S-RoBERTa finetuned on NLI data
as well as Universal Sentence Encoder
(USE) \citep{cer-etal-2018-universal} and
InferSent \citep{conneau-etal-2017-supervised}, all of which
are trained on hundreds of thousands of labeled text pairs
from SNLI \citep{bowman-etal-2015-large} and
MNLI \citep{williams2018mnli}. We additionally compare to
the following fully unsupervised approaches: averaging word-level GloVe \citep{Glove} or BERT \citep{devlin2018bert} embeddings, using BERT's CLS token, and recent methods by \citet{zhang-etal-2020-unsupervised} and \citet{li-etal-2020-sentence} based on pretrained BERT models. We do not compare to approaches trained with direct supervision as our focus is on obtaining sentence representations without task-specific labeled examples.
As shown in Table~\ref{table:main-results}, training on datasets generated with \ours{} clearly outperforms the fully unsupervised baselines; on average, training on \ourDsSemi{} even outperforms all approaches with NLI supervision. 
\ourDsSemi{} gives better results than \ourDsFull{} on all STS datasets as its examples are -- by design -- very similar to examples found in these datasets, while the latter gives better results on SICK.

We investigate the importance of self-debiasing \citep{schick2021selfdiagnosis} in Table~\ref{table:self-debiasing} (top); as can be seen, removing self-debiasing ($\lambda = 0$) dramatically hurts performance.
Increasing the decay constant ($\lambda = 200$) leads to slightly worse performance as the overall quality of generated sentences decreases \citep{schick2021selfdiagnosis}.
Table~\ref{table:self-debiasing} (bottom) shows that training on \ourDs{} requires measures to limit the effect of noisy labels: removing label smoothing and performing no data augmentation (i.e., not generating additional pairs $(\mathbf{x}_1, \mathbf{x}_2, 0)$ by sampling random $\mathbf{x}_2$'s for each $\mathbf{x}_1$) clearly hurts performance. 

To further assess the quality of datasets generated with \ours{}, we additionally perform a small-scale human evaluation. To this end, we consider the exact version of \ourDsSemi{} used for training S-RoBERTa; that is, we perform label smoothing, augmentation with randomly sampled text pairs, and removal of trivial examples where $\mathbf{x}_1\,{=}\,\mathbf{x}_2$. From the resulting dataset, we randomly select 100 text pairs $(\mathbf{x}_1,\mathbf{x}_2)$ and annotate them ourselves with similarity scores $y \in \{0, 0.1, 0.5, 0.9\}$, where we assign a score of $0.9$ when $\mathbf{x}_1$ and $\mathbf{x}_2$ mean (almost) the same thing and a score of $0.1$ when they are on different topics, but still show a weak similarity in some aspect.

In Table~\ref{table:human-eval}, human annotations are compared to originally assigned scores, yielding some interesting insights. For one, it becomes clear why augmentation with randomly sampled text pairs is important for good downstream task performance: Of the examples generated by \ours{} that are supposed to be on completely different topics, many (41\%) still have a certain similarity according to human judgment. In contrast, randomly sampled pairs are indeed on completely different topics in almost all cases. Moreover, we can see that GPT2-XL has particular difficulty in generating pairs of non-identical sentences that really mean the same thing: Only 47\% of all examples that should have the same meaning do actually mean (almost) the same thing. However, the strong performance of S-RoBERTa trained on \ourDsSemi{} suggests that, despite this noise, there is sufficient signal in this dataset for successful training.

\begin{table}
	\small
	\centering
	{
		\def\arraystretch{1.5}%
		\setlength\tabcolsep{2pt}
		\begin{tabular}{rrcccccccc}
			\multicolumn{2}{r}{\ours{} Labels $\rightarrow$} & $\mathsf{0.0}$ && $\mathsf{0.1}$ && $\mathsf{0.5}$ &&  $\mathsf{0.9}$ & \\
			\parbox[t]{0mm}{\multirow{4}{*}{\hskip25pt\rotatebox[origin=c]{90}{{Human Labels}}}} 
			&\qquad$\mathsf{0.0}$\ \ {} & \cca{95}{\quad{\bt95\%}\quad}  && \cca{15}{\quad15\%\quad} 		&& \cca{1}{\quad\phantom{0}0\%\quad} 		&& \cca{1}{\quad\phantom{0}0\%\quad} & \\
			& $\mathsf{0.1}$\ \ {} & \cca{1}{\phantom{0}0\%} 		&& \cca{44}{\bt 44\%} 	&& \cca{11}{11\%} 		&& \cca{12}{12\%} & \\
			& $\mathsf{0.5}$\ \ {} & \cca{5}{\phantom{0}5\%} 		&& \cca{41}{41\%} 	&& \cca{60}{\bt 60\%} 	&& \cca{41}{41\%} & \\
			& $\mathsf{0.9}$\ \ {} & \cca{1}{\phantom{0}0\%} 		&& \cca{1}{\phantom{0}0\%} 		&& \cca{29}{29\%} 	&& \cca{47}{\bt 47\%} & \\
	\end{tabular}}
	\caption{Comparison of similarity scores in \ourDsSemi{} to human judgments for 100 examples. Examples are chosen randomly from the version of \ourDsSemi{} used for training (including label smoothing, augmentation with random pairs and removal of examples where $\mathbf{x}_1 = \mathbf{x}_2$). For column $i$ and row $j$, the value shown is the percentage of examples generated by \ours{} for similarity score $i$ that were assigned score $j$ in our human evaluation.}
	\label{table:human-eval}
\end{table}

We finally take a qualitative look at both positive examples where \ours{} is able to create high-quality text pairs and at some typical errors found in many of the generated examples. As shown in Table~\ref{table:qualitative-analysis}, for $y=1$ the PLM sometimes comes up with decent paraphrases (e.g. ``notches a victory'' $\mapsto$ ``wins'') or substitutes with very similar meaning (``cutting'' $\mapsto$ ``slicing''), but more often it generates sentences that either omit or mix up important information, and sometimes it produces sentences with an entirely different meaning. Whereas sentences generated for $y=0.5$ by and large look reasonable, for $y=0$ the PLM often simply flips words (``closed'' $\mapsto$ ``open'', ``large'' $\mapsto$ ``small'') instead of producing sentences on completely different topics.

\begin{table}
	\small \setlength\tabcolsep{3pt} 
	\begin{tabularx}{\linewidth}{cXc}
		 \toprule
		\arrayrulecolor{decentgrey}
		\parbox[t]{3mm}{\multirow{10}{*}{\rotatebox[origin=c]{90}{{\hskip-0.8cm$y=1$}}}} 
		& $\mathbf{x}_1\,{=}\,$ \lsstyle Rick{\,}Santorum{\,}notches\,a\,victory\,in\,Kansas\,caucuses. & \multirow{2}{*}{\cmark} \\
		& $\mathbf{x}_2\,{=}\,$ \lsstyle Rick Santorum wins Kansas caucuses. & \\ 
		\cmidrule{2-3}
		& $\mathbf{x}_1\,{=}\,$ A man is cutting cucumbers. & \multirow{2}{*}{\cmark} \\
		& $\mathbf{x}_2\,{=}\,$ A man is slicing cucumbers. & \\
		\cmidrule{2-3}
		& $\mathbf{x}_1\,{=}\,$ US closes embassy in Syria & \multirow{2}{*}{\xmark} \\
		& $\mathbf{x}_2\,{=}\,$ US Embassy in Syria & \\
		\cmidrule{2-3}
		& $\mathbf{x}_1\,{=}\,$ A man is playing the cello. & \multirow{2}{*}{\xmark} \\
		& $\mathbf{x}_2\,{=}\,$ The cello is playing the man. & \\
		\cmidrule{2-3}
		& $\mathbf{x}_1\,{=}\,$ A plane is taking off. & \multirow{2}{*}{\xmark} \\
		& $\mathbf{x}_2\,{=}\,$ I want to be a pilot. & \\
		\arrayrulecolor{black}
		\midrule 
		\arrayrulecolor{decentgrey}
		\parbox[t]{3mm}{\multirow{4}{*}{\rotatebox[origin=c]{90}{{\hskip-0.2cm$y=0.5$}}}}
		& $\mathbf{x}_1\,{=}\,$ \lsstyle A woman is seasoning a piece of meat. & \multirow{2}{*}{\cmark} \\
		& $\mathbf{x}_2\,{=}\,$ \lsstyle A man is cooking the meat and adding spices [...] & \\ 
		\cmidrule{2-3}
		& $\mathbf{x}_1\,{=}\,$ \lsstyle Second day of Egyptian presidential election & \multirow{2}{*}{\cmark} \\
		& $\mathbf{x}_2\,{=}\,$ The first night of the election. & \\	
		\arrayrulecolor{black}
		\midrule 
		\arrayrulecolor{decentgrey}
		\parbox[t]{3mm}{\multirow{10}{*}{\rotatebox[origin=c]{90}{{\hskip-0.8cm$y=0$}}}}
		& $\mathbf{x}_1\,{=}\,$ \lsstyle A white bus with the word Julia is near water [...] & \multirow{2}{*}{\cmark} \\
		& $\mathbf{x}_2\,{=}\,$ \lsstyle There is an open beach in my hometown. & \\ 
		\cmidrule{2-3}
		& $\mathbf{x}_1\,{=}\,$ Strong earthquake in Mexico & \multirow{2}{*}{\cmark} \\
		& $\mathbf{x}_2\,{=}\,$ It's the best time to get a job & \\
		\cmidrule{2-3}
		& $\mathbf{x}_1\,{=}\,$ Closed roads in Armenia & \multirow{2}{*}{\xmark} \\
		& $\mathbf{x}_2\,{=}\,$ Open roads in Azerbaijan & \\
		\cmidrule{2-3}
		& $\mathbf{x}_1\,{=}\,$ The man is playing the guitar. & \multirow{2}{*}{\xmark} \\
		& $\mathbf{x}_2\,{=}\,$ I'm not a guitar player. & \\
		\cmidrule{2-3}
		& $\mathbf{x}_1\,{=}\,$ A man is playing a large flute. & \multirow{2}{*}{\xmark} \\
		& $\mathbf{x}_2\,{=}\,$ A man is listening to a small flute. & \\
		\arrayrulecolor{black}
		\bottomrule
	\end{tabularx} 
	\caption{A selection of high-quality (\cmark) and low-quality (\xmark) examples in \ourDsSemi{}. Many sentence pairs for $y=1$ are not similar and have quite different meanings. Some sentence pairs for $y=0$ are not on completely different topics.}  
	\label{table:qualitative-analysis}
\end{table}

\section{Conclusion}

We have introduced \ours{}, a method for using large PLMs to generate entire datasets of labeled sentence pairs from scratch, requiring no labeled data and no parameter updates. This is achieved by providing instructions in natural language, combined with the self-debiasing method of \citet{schick2021selfdiagnosis}. With appropriate measures for handling noisy data, models trained on datasets generated with \ours{} achieve strong results on several semantic textual similarity datasets. 

For future work, it would be interesting to see whether the noise in datasets generated with \ours{} can further be reduced, e.g., by using different sets of instructions \citep{jiang2019know,schick2020exploiting} or by supplementing our pipeline with some additional filtering steps.

\paragraph*{Acknowledgments}
This work was funded by the European Research Council (ERC \#740516).
We thank the anonymous reviewers
for their helpful comments.

\bibliography{literatur}
\bibliographystyle{acl_natbib}

\clearpage
\appendix
\section{Experimental Setup}

Our implementation is based on the Transformers library \citep{wolf2019transformers} and PyTorch \citep{paszke2017automatic}. All our experiments were conducted using two GPUs with 11GB RAM (NVIDIA GeForce GTX 1080 Ti). Generating \ourDsFull{} and \ourDsSemi{} using both GPUs took approximately 48 hours per dataset. Training a Sentence Transformer on these datasets took less than 2 hours on average.

\section{Datasets}

Both datasets generated with \ours{} (\ourDsFull{} and \ourDsSemi{}) are publicly available at \url{https://github.com/timoschick/dino}. After filtering out examples where the language model did not produce a quotation mark, \ourDsSemi{} contains 121,275 examples and \ourDsFull{} contains 143,968 examples.

\section{Additional Results}

Our main results do not include scores for DeCLUTR \citep{giorgi2020declutr} and CLEAR \citep{wu2020clear} -- two recent approaches using contrastive learning -- as their evaluation setup differs from that described in \citet{reimers-gurevych-2019-sentence} (and used by all other baselines) in the following respects:
\begin{itemize}
	\item Both \citet{giorgi2020declutr} and \citet{wu2020clear} treat SICK and STSb as \emph{supervised} tasks, i.e., they use the provided task-specific training sets to perform regular supervised training.
	\item The STS12--16 datasets each consist of several subsets. \citet{giorgi2020declutr} and \citet{wu2020clear} compute Spearman's correlation coefficient separately for each of these subsets and report the mean score across all subsets. In contrast, for our main results we follow \citet{reimers-gurevych-2019-sentence} and concatenate all subsets to form one large set on which Spearman's correlation is computed just once.
\end{itemize}
As the implementations of both methods are not publicly available as of this writing, we are unable to compute scores for DeCLUTR and CLEAR using the evaluation setup of \citet{reimers-gurevych-2019-sentence} ourselves. Instead, we recompute scores for \ours{} (both with \ourDsSemi{} and \ourDsFull{}) using the evaluation setup of \citet{giorgi2020declutr} and \citet{wu2020clear} on STS12--16; results are shown in Table~\ref{table:additional-results}.

\begin{table}[h!]
	\small
	\setlength\tabcolsep{2pt}
	\lsstyle
	\renewcommand{\arraystretch}{0.92}
	\begin{tabularx}{\linewidth}{lYYYYYY}
		\toprule
		\textbf{Model} & \textbf{STS12} & \textbf{STS13} & \textbf{STS14} & \textbf{STS15} & \textbf{STS16} & \textbf{Avg.} \\
		\midrule
		CLEAR     & 49.0 & 48.9 & 57.4 & 63.6 & 65.6 & 56.9 \\
		DeCLUTR   & 64.2 & 70.4 & 70.0 & \textbf{77.5} & 75.4 & 71.5 \\
		\ourDsFull{}      & 65.1 & 69.9 & 68.6 & 76.3 & 76.6 & 71.3 \\
		\ourDsSemi{}   & \textbf{65.3} & \textbf{71.8} & \textbf{72.7} & 75.9 & \textbf{76.9} & \textbf{72.5} \\
		\bottomrule
	\end{tabularx}
	\caption{Results for CLEAR \citep{wu2020clear}, DeCLUTR \citep{giorgi2020declutr} and Sentence-RoBERTa (base) trained on \ourDsFull{} and \ourDsSemi{} using the evaluation setup of \citet{wu2020clear} and \citet{giorgi2020declutr}: For each task, we report the mean Spearman correlation of all subtasks in a fully unsupervised setting.}
	\label{table:additional-results}
\end{table}

\end{document}